\documentclass{article}

\usepackage{PRIMEarxiv}

\usepackage[utf8]{inputenc} 
\usepackage[T1]{fontenc}    
\usepackage[colorlinks=true, pdfborder={0 0 0}]{hyperref}
\usepackage{url}            
\usepackage{booktabs}       
\usepackage{amsfonts}       
\usepackage{amsmath}        
\usepackage{nicefrac}       
\usepackage{microtype}      
\usepackage{adjustbox}
\usepackage{fancyhdr}       
\usepackage{graphicx}       
\graphicspath{{media/}}     
\usepackage{xcolor}

\title{Similarity-based Feature Extraction for Large-scale Sparse Traffic Forecasting}

\author{
  Xinhua Wu\thanks{The authors contributed equally to this work.}  \\
  Department of Civil and Environmental Engineering \\
  Northeastern University\\
  Boston, MA, USA\\
  \texttt{wu.xinh@northeastern.edu} \\
   \And
  Cheng Lyu\footnotemark[1], Qing-Long Lu\footnotemark[1], and Vishal Mahajan \\
  Chair of Transportation Systems Engineering\\
  TUM School of Engineering and Design \\
  Technical University of Munich \\
  Munich, Germany \\
  \texttt{\{cheng.lyu, qinglong.lu, vishal.mahajan\}@tum.de} \\
}

\begin{document}
\maketitle

\begin{abstract}
  Short-term traffic forecasting is an extensively studied topic in the field of intelligent transportation system. However, most existing forecasting systems are limited by the requirement of real-time probe vehicle data because of their formulation as a time series forecasting problem. Towards this issue, the NeurIPS 2022 Traffic4cast challenge is dedicated to predicting the citywide traffic states with publicly available sparse loop count data. This technical report introduces our second-place winning solution to the extended challenge of ETA prediction. We present a similarity-based feature extraction method using multiple nearest neighbor (NN) filters. Similarity-based features, static features, node flow features and combined features of segments are extracted for training the gradient boosting decision tree model. Experimental results on three cities (including London, Madrid and Melbourne) demonstrate the strong predictive performance of our approach, which outperforms a number of graph-neural-network-based solutions in the task of travel time estimation. The source code is available at \url{https://github.com/c-lyu/Traffic4Cast2022-TSE}.
\end{abstract}


\setcounter{footnote}{0} 

\section{Introduction}
Traffic forecasting is a crucial task in intelligent transportation systems. Short-term traffic forecasting models provide transport service operators with future traffic conditions to support more accurate real-time routing information and better congestion mitigation measures. Traffic congestion is a spatio-temporal phenomenon borne out of complex interactions of traffic demand (activity location, mode choice, destination choice, route choice, etc.) with the transport supply (transport networks, real-time disruptions, etc.) \cite{JRC69961}. Thus, the availability of rich data capturing spatial, temporal, and network correlations is essential to predict future traffic conditions.

The \textit{Traffic4cast} competition series hosted by the Institute of Advanced Research in Artificial Intelligence (IARAI) is dedicated to benchmarking existing traffic forecasting systems and to advancing the state-of-the-art. The competition data is based on $10^{12}$ data points and for the first time allows the scientific community to study the dynamic link between loop counter data time series and the traffic state dynamics of entire cities rather than small, non-isolated traffic subsystems. 

Beyond the success of \textit{Traffic4cast} 2019, 2020 and 2021, \textit{Traffic4cast} 2022 poses a new challenge. Participants are invited to make traffic state predictions  on road graph edges from sparse loop counter data. While accurate sensing and prediction of citywide traffic state using only sparse loop counter data is challenging, it can significantly lower the technological and financial barrier of entry. Lower data requirements and lighter models enable the solution to be more easily implemented and deployed.

In this technical report, we present a feature extraction method based on traffic state similarity and build gradient boosting tree models for prediction.


\section{Problem Statement}
\label{sec:headings}

\subsection{Loop count data}
Loop detectors have been widely deployed in urban areas for effective traffic surveillance since the 1980s \cite{coifman2001improved, klein2006traffic}. A loop detector, buried under the pavement, can detect a vehicle passing over or arriving at a certain point, for instance approaching a signalized intersection \cite{kwon2003estimation}. The traffic volume information can thus be derived by aggregating the loop detector's readings over a certain time period (typically aggregated every 15 min) \cite{meng2017city}.

Although loop detectors can accurately count traffic volumes, two important drawbacks prevent us from getting an accurate sense of the citywide traffic dynamics from the loop count data. First, the data is spatially sparse. Since the implementations of loop detectors are expensive, the government can only implement them on a small number of critical road segments in practice. 
For example, in the three studied cities, the number of loop counters is under 4,000 in a city network with over 100,000 edges. Secondly, aged loop detectors may frequently experience equipment failures, resulting in missing data in a relatively long period. 

\subsection{Probe vehicle data}
GPS data from probe vehicles have been getting more attention lately. Taxis, buses and private vehicles with activated navigation devices can serve as probe vehicles in urban traffic systems. The spatio-temporally aggregated information derived from probe vehicle trajectories, including volume, speed and travel time, can represent the traffic condition on each segment in a certain time period \cite{hunter2009path}.

However, probe vehicle data also has its limitations. Notably, GPS fleets (specifically probe vehicles from HERE Technologies in this competition) comprise only a subset of all vehicles in traffic. Segments, especially those with low traffic volumes at non-peak periods, may have insufficient probe vehicles for the inference of their traffic conditions at some time periods \cite{zheng2013urban}. In other words, probe vehicle data are temporally sparse. In addition, raw probe vehicle data is large and complex compared to the loop count data. Real-time citywide traffic state predictions with probe vehicle data are technically and financially difficult in production systems, especially when a highly complicated inference model is deployed \cite{heit2016architecture}.

\subsection{Objective}
The objective of this competition is to predict future traffic states with merely loop count data. The underlying assumption is that the loop count data is a good encoder to represent much of the complex traffic dynamics. The sparsity of loop detector data and the lack of long-term time series data increase the challenge of the task. On the other hand, this also gives us a chance to build lighter and more deployable predictive models which are based on easily available public loop data.
The competition is comprised of two tasks: the core challenge (traffic state prediction) and the extended challenge (travel time prediction).
In this report, we focus on the latter, which aims to accurately predict the future travel time of supersegments in the next 15 minutes with sparse loop count data. 

The static road network graph of a city is denoted by $\mathcal{G}(\mathcal{V}, \mathcal{E}, \mathcal{S})$. Herein, $\mathcal{V} = \{v_i\} \in \mathbb{R}^{|\mathcal{V}|}$ is the node set, where the number of nodes is represented by $|\mathcal{V}|$. $\mathcal{E} = \{e_i = (v_j, v_k) | v_j, v_k \in \mathcal{V}, v_j \neq v_k \} \in \mathbb{R}^{|\mathcal{E}|}$ is the edge set, and the number of edges is given by $|\mathcal{E}|$. 
Many paths with indefinite lengths (referred as supersegments) are sampled from the graph to form the supersegment set $\mathcal{S} = \{(e_i, e_j, \dots)\}$, and we denote the number of supersegments as $|\mathcal{S}|$.

We also denote the node flow vector at a specific time step $t$ by $\mathcal{X}^t \in \mathbb{R}^{|\mathcal{V}|}$, the $i$-th element of which is the 15-minute traffic volume of the $i$-th loop counter. Similarly, the travel time vector of all supersegments at time step $t$ is denoted by $\mathcal{Y}^t \in \mathbb{R}^{|\mathcal{S}|}$.

\paragraph{Problem: Supersegment travel time prediction.}
Given $4$ historical node flow vectors, we want to predict the travel time of all supersegments in the following time step. A function $f(\cdot)$ is expected to be learned with the static road network graph $\mathcal{G}$ in addition to historical node flow:
\begin{equation}
    f: \left[ \mathcal{X}^{t-4}, \mathcal{X}^{t-3}, \mathcal{X}^{t-2}, \mathcal{X}^{t-1} ; \mathcal{G}\right]
    \to \mathcal{Y}^t
\end{equation}


\section{Method}
\label{sec:others}

\subsection{Framework overview}

\begin{figure}[hbtp]
    \centering
    \includegraphics[width=\columnwidth]{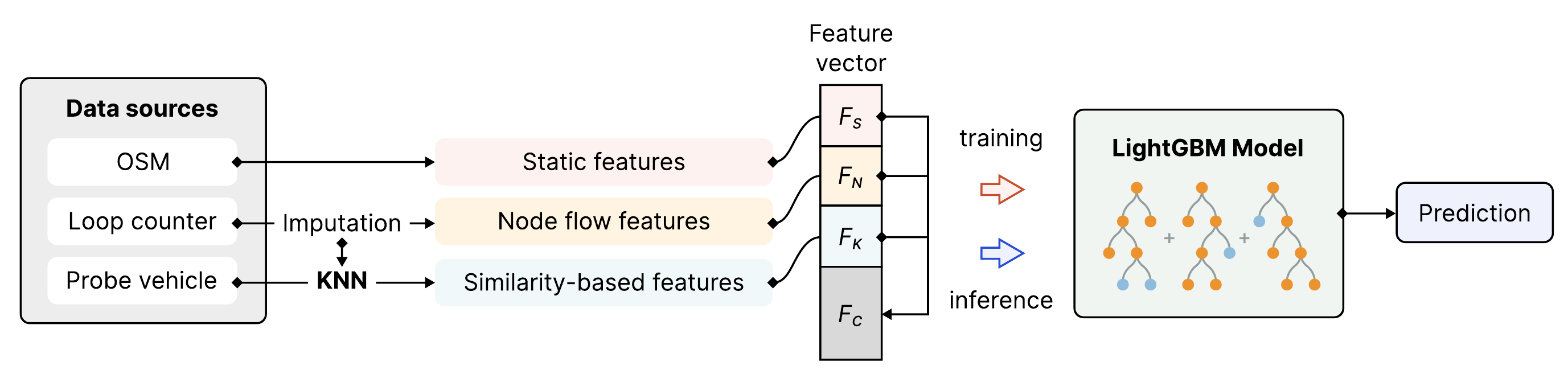}
    \caption{Overall framework of the proposed method.}
    \label{fig:overview}
\end{figure}

An overview of the proposed method is shown in Figure~\ref{fig:overview}.  Various features are extracted from three data sources, and then fed into the LightGBM model for training and inference.

During feature extraction on the training data, data imputation is used to deal with the temporal missing values of loop count data. Note that the imputation is feasible only on the training set as the full time series of the test set is unavailable. Details pertaining to data imputation can be found in Section \ref{sec:data imputation}. 

For each supersegment at a specific time step, four groups of features are extracted, including static features $F_S$, node flow features $F_N$, similarity-based features $F_K$ and combined features $F_C$. Notably, the proposed similarity-based feature extraction significantly increases the model performance. Details pertaining to feature extraction can be found in Section \ref{sec:feature extraction}. 

\subsection{Data imputation}
\label{sec:data imputation}
A large amount of missing data exists in the traffic volume collected by loop counters (i.e., the node data), both spatially and temporally. The data missing in the spatial domain can be mainly attributed to the lack of loop counters, but it is not the focus of this section. The temporal absence of data can be either spontaneous or planned, where the spontaneous ones are possibly caused by unexpected malfunctioning of the sensors, while the planned ones may be due to system maintenance. Among the two types of causes, it is inferred that the planned missing accounts for the majority of missing as most of the gaps in the dataset show up on a regular basis or last for a relatively long time over multiple neighboring nodes. Therefore, we manage to address data imputation with the assumption of independence between nodes, given that neighboring nodes do not provide much valuable information for restoring missing values.

We model the traffic volume collected by a loop counter as a univariate Gaussian process with a periodic prior. Denote the index of time $i$ as $x_i = i$ and the corresponding traffic volume as $y_i$. We express the traffic volume as a function of time, $y_i = f(x_i) + \varepsilon_i$, with a zero-mean Gaussian noise $\varepsilon_i$. Then, all traffic volume values along the time domain $\{y_1, y_2, \dots, y_T\}$ are assumed to together follow a multivariate Gaussian prior $\boldsymbol{Y} \sim \mathcal{N}(\boldsymbol{\mu}, \Sigma)$, where the mean of an arbitrary input $x_i$ and the covariance of two arbitrary inputs $x_i$ and $x_j$ are characterized by a mean function $m(x_i)$ and a covariance function $\kappa(x_i, x_j)$ respectively.

To realize our goal of data imputation, we are interested in drawing the posterior distribution at missing points based on known values of the node traffic volumes. Without loss of generality, assuming the mean function values as constant zeros, the posterior given test input $x_\star$ can then be obtained by conditioning the joint prior distribution on training samples $\{\boldsymbol{x}, \boldsymbol{y}\}$ \cite{rasmussen2006gpml}, 
\begin{align}
\bar{y}(x_\star) &= \kappa'(x_\star, \boldsymbol{x}) \Sigma^{-1} \boldsymbol{y},\\
\sigma^2_y(x_\star) &= \kappa(x_\star, x_\star) - \kappa'(x_\star, \boldsymbol{x}) \Sigma^{-1} \kappa(x_\star, \boldsymbol{x}).
\end{align}

An appropriate covariance function is required to effectively restore the periodicity of traffic volumes. The basic periodic function used in this report is formulated as,
\begin{equation}
\kappa(x_i, x_j;P,l) = \exp\left( -2l^{-2} \sin^2\left( \frac{\pi}{P}|x_i - x_j| \right) \right).
\end{equation}
where $P$ denotes the periodicity of the kernel and $l$ denotes the length scale.

Instead of modelling the traffic volume of the entire time range, we address a relatively small segment once at a time in order to capture the local patterns of traffic dynamics. To further incorporate the local trend of time series, we combine the basic periodic function with the dot product function ($\kappa(x_i, x_j) = x_i x_j$) as,
\begin{equation}
\begin{aligned}
\kappa(x_i, x_j;P_1,l_1,P_2,l_2) =& \exp\left( -2{l_1}^{-2} \sin^2\left( \frac{\pi}{P_1}|x_i - x_j| \right) \right) x_i x_j\\
&+ \exp\left( -2{l_2}^{-2} \sin^2\left( \frac{\pi}{P_2}|x_i - x_j| \right) \right),
\end{aligned}
\label{eq:gp-kernel}
\end{equation}
where the first periodic function, together with the dot product function, attends to daily periodicity, whilst the second attends to subtle fluctuation. 

\subsection{Feature extraction}
\label{sec:feature extraction}

Features are extracted from the datasets to support the training of the gradient boosting model. These features can be classified into four groups, which include similarity-based features, static features, node flow features and combined features.

\paragraph{Similarity-based features}
The similarity-based feature extraction is inspired by the assumption that loop count data are good encoders of the citywide traffic state. Ideally, the temporary neighbors of a time period (i.e, these time periods with similar loop count data) should contain most relevant traffic state information.

\begin{figure}
    \centering
    \includegraphics[width=0.7\columnwidth]{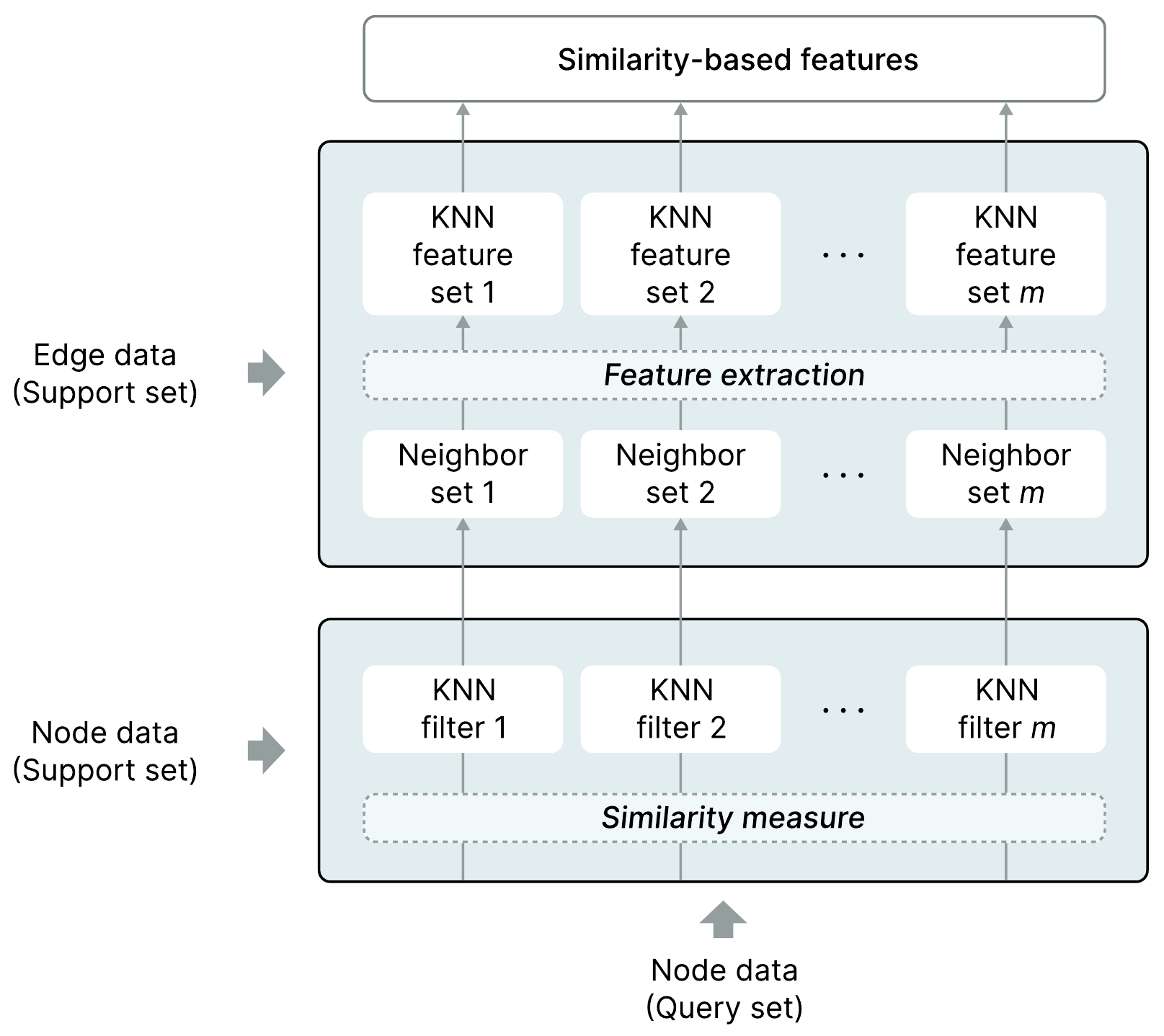}
    \caption{Extracting similarity-based features using KNN.}
    \label{fig:knn}
\end{figure}

The similarity-based feature extraction is shown in Figure \ref{fig:knn}. Here, the query set refers to these for which we expect to find neighbors, while the support set is the according candidate set for neighbors of the query set. Note that two time periods that are temporally close or overlap may have highly correlated node flows and traffic states, which can lead to serious data leakage. Therefore, one principle for splitting the query set and the support set is to make sure that they are not temporally close. Following this principle, two strategies were used to split the training set. 1) \textbf{Equal split}. In the training set, we withdraw one week of data from every two weeks as the query set while the rest is served as the support set; 2) \textbf{Day-wise splitting}. The data from each individual day are used as the query set and the rest of the training set are used as the support set accordingly.

Using a specific KNN filter, a neighbor set of every time period in the query set can be found based on the similarity of node data (i.e., loop count). A total of $m$ KNN filters are introduced to better capture the similarity. These KNN filters differ in terms of the value of $k$ and the measure of distance. Furthermore, aggregated statistical features can be extracted from $m$ neighbor sets respectively, such as mean and standard deviation of historical travel time. The collection of these feature sets from different KNN filters constitutes the complete set of similarity-based features.

\paragraph{Static features}
Network-based features are those extracted from the OSM data describing the characteristics of edges and supersegments, that restrain the potential label range of the respective edges and supersegments. For instance, speed limits on the edges would determine the shortest travel time for traversing it and further influence the supersegment travel time. To further capture supersegments properties, we also statistically summarized their historical labels (all-NN).

\paragraph{Node flow features}
Node flow features are devised using the loop count data. Since a supersegment is defined as a path connecting a group of nodes, we believe that aggregating the information captured by the loops installed at nodes can shed some light on the traffic state along the supersegment.
 
\paragraph{Combined features}
Combined features are designed with some of the above features, for example, the ratio between the median travel time from a 30-NN filter and a 100-NN filter may further represent the variety of the supersegment travel time.


\section{Experiment Results}

\subsection{Experimental setup}
\paragraph{Dataset description} 
We evaluated the performance of the proposed method on the loop count data and probe vehicle data\footnote{Provided by HERE Technologies, \url{https://developer.here.com/sample-data}.} of three cities, namely London, Madrid and Melbourne. Brief statistics of the data are provided in Table~\ref{tab:dataset}. The three datasets share similar scales in terms of the number of days, nodes, edges and supersegments. However, because of the discrepancy in urban structure and traffic characteristics, there is an apparent inconsistency among them with regard to the distribution of labels, including both edge congestion classes and supersegment ETAs.

\begin{table}[hbtp]
\centering
\caption{Summary statistics of dataset.}
\label{tab:dataset}
\begin{tabular}{lrrr} 
    \toprule
      & \textbf{ London } & \textbf{ Madrid } & \textbf{ Melbourne } \\ 
    \midrule
    \multicolumn{4}{c}{\textit{Basic information}} \\ 
    \midrule
    Start date & 2019/07/01 & 2021/06/01 & 2020/06/01 \\
    End date & 2020/01/31 & 2021/12/31 & 2020/12/30 \\
    \# of days & 110 & 109 & 106 \\
    \# of nodes & 59110 & 63397 & 49510 \\
    \# of counters & 3751 & 3875 & 3982 \\
    \# of edges & 132414 & 121902 & 94871 \\
    \# of supersegments & 4012 & 3969 & 3246 \\ 
    \midrule
    \multicolumn{4}{c}{\textit{Supersegment ETA}} \\ 
    \midrule
    Min (s) & 16.2 & 15.7 & 19.1 \\
    Mean (s) & 393.3 & 222.2 & 343.8 \\
    Max (s) & 3600 & 3600 & 3600 \\
    \bottomrule
\end{tabular}
\end{table}

\paragraph{Evaluation metric} 

As a regression problem, in the extended challenge of travel time estimation, L1 loss, a.k.a., mean absolute error, is used to measure the prediction error. 
\begin{equation}
\ell(\hat{y}, y) = \frac{1}{N} \sum_{n=1}^N |\hat{y}_n - y_n|
\end{equation}

\subsection{Results and discussion}

We experimented with different KNN similarity measures, data splitting strategies and feature groups to assess the effectiveness of the techniques involved in the proposed method. Table~\ref{tab:model_comparison} compares the model performance under different experimental setups. The optimal configuration leads to an improvement of 4.1\% in MAE (from 62.3360 to 59.7973) compared to the baseline model without considering the noise introduced by these aspects.
\begin{table}[hbtp]
\centering
\caption{Model comparison.}
\label{tab:model_comparison}
\begin{adjustbox}{center}
\begin{tabular}{llllll}
    \toprule
    \textbf{ Splitting } &\textbf{ Distance (speed) } & \textbf{ Distance ($y$) } & \textbf{ \# of trees } & \textbf{ \# of leaves } & \textbf{ MAE } \\ 
    \midrule
    Equal & Euclidean & Euclidean & 1500 & 42 & 62.3360 \\
    Day-wise & Euclidean & Euclidean & 1500 & 42 & 61.7966 \\
    Day-wise & Manhattan & Euclidean & 1500 & 42 & 59.9625 \\
    Day-wise & Manhattan & Manhattan & 1500 & 42 & 59.9174 \\
    Day-wise & Manhattan & Euclidean \& Manhattan & 1500 & 42 & 59.9139 \\
    Day-wise & Manhattan & Euclidean \& Manhattan \& nor. Euc. & 1500 & 42 & 59.9142 \\
    Day-wise & Manhattan & Euclidean \& Manhattan & 2000 & 64 & 59.7973 \\
    Day-wise & Manhattan & Euclidean \& nor. Euc. & 2000 & 64 & 59.8096 \\
    \bottomrule
\end{tabular}
\end{adjustbox}
\end{table}

\paragraph{Similarity measure}
Three different similarity measures were applied and compared in the experiments, namely, Manhattan distance, Euclidean distance, and normalized Euclidean distance, where the normalized Euclidean distance is computed on the node-wise normalized loop data. The results present the superiority of Manhattan distance in the extended challenge. More specifically, changing the similarity measure from Euclidean distance to Manhattan distance reduces the L1 loss by 3.1\% (from 61.8 to 59.9). One potential reason is, Euclidean distance imposes great penalties on the nodes with opposite missing statuses in the two samples being compared. Due to the high missing rate in the loop count data, the accumulative penalty will be a large number and dominate the neighbor search procedure. Recall that the temporal absence of data is primarily caused by sensors malfunctioning, so the model should skip these nodes when computing the distance rather than treating them as anomalies. Despite the nodes with missing values differ from time step to time step preventing their removal, applying Manhattan distance to measure the similarity can mitigate the influence of missing values on neighbor selection. This conjecture has also been supported by the close losses resulting from the model using Manhattan distance-based KNN features and the one using normalized Euclidean distance-based KNN features. Further, we found that combining the similarity-based features generated by the KNN filters equipped with different similarity measures in model training can result in a smaller loss. 


\paragraph{Dataset splitting}
The reliability of KNN highly depends on the quantity and quality of the trainset. The dataset splitting method thus has a critical impact on the quality of the neighbors selected. Recall that two splitting methods have been tested in the experiments, i.e., equal and day-wise splitting. The results show that, while the other conditions hold constant, the latter method reduces the loss by 0.8\% (from 62.3 to 61.8). It is worth mentioning that, for the samples with high missing rates (about 90\% missing) in the London case, their neighbors are retrieved from the rest samples subject to the nodes with valid values. These samples account for about 6\% of the entire London dataset, and a special treatment helps decrease the loss by 2.1 for London (i.e., 0.7 for three city average).

\paragraph{Feature importance}
A variety of feature sets were examined to search for the most useful features. The feature importance is evaluated by
the total of information gain resulting from the corresponding splits of each feature. 
As a reference, Table~\ref{tab:feature-importance} lists the top-20 most important features for the model fitted with London data (normalized by the total information gain of all features). 
Among all groups of features, \textit{similarity-based features}, as well as the derived \textit{combined features}, contribute the most to the prediction result. It shows that the proposed similarity filtering method can effectively capture the global urban traffic state. It is also worth noting that supersegment ID is the only feature unrelated with the KNN feature extractor, implying the heterogeneity between supersegments.


\section{Conclusion}
In this report, we present our second-place winning solution to \textit{Traffic4cast} 2022. Experiment results have verified the possibility of non-deep models in generalizing sparse node observations of traffic states to the entire road network. Notably, the proposed similarity-based feature extraction method is capable of effectively identifying the underlying global traffic patterns utilizing multiple KNN filters with various similarity measures. As a global feature extractor, it is unaffected by the spatial sparsity of node data, which, however, also becomes a limit to the proposed method in terms of the ability of capturing localized features. For future work, more emphasis can be placed on the design of a local feature extractor, so as to improve its predictive power in both global and local aspects.


\bibliographystyle{unsrt}  
\bibliography{t4c22}  

\appendix
\section*{Appendix}
\section{Parameter settings}
\subsection{Data imputation}
The Gaussian process model used for data imputation is fully characterized by the prior covariance function. The basic parameters, as shown in Equation~\eqref{eq:gp-kernel}, include the periodicity and length-scales of the two periodic function, i.e., $P_1, l_1, P_2, l_2$. Additionally, we add a white noise $\epsilon$ to the diagonal of covariance kernel $\Sigma$ to allow for noises in the training data. These parameters can be optimized through marginal likelihood maximization. During optimization, we imposed feasible ranges (see Table~\ref{tab:gp-param}) for length-scales and the white noise, while keeping the periodicity constant, to encode our prior of the volume pattern.

\begin{table}[hbtp]
\centering
\caption{Prior range of Gaussian process parameters.}
\label{tab:gp-param}
\begin{tabular}{cr} 
\toprule
Parameter & Prior range \\ 
\midrule
$P_1$ & $95$ \\
$l_1$ & $[100, 1000]$ \\
$P_2$ & $95$ \\
$l_2$ & $[0.5, 1]$ \\
$\epsilon$ & $[0.001, 1]$ \\
\bottomrule
\end{tabular}
\end{table}

\subsection{Model training}
The LightGBM models trained for this competition (the extended challenge) adopted the same loss function as the evaluation metric, i.e., L1 loss. Apart from the number of trees for boosting, we also specified learning rate, the maximum tree depth, the number of leaves during training. Additionally, we subsampled the features instead of adding regularization terms to alleviate overfitting. A list of major hyperparameters of the final model is shown in Table~\ref{tab:lgb-param}. Experiments were conducted on an Intel® Xeon(R) Gold 6240 CPU @ 2.60GHz × 72 with 156.9 GiB RAM. It takes about 30 minutes to train the model for a city with the hyperparameters specified in Table~\ref{tab:lgb-param}. During the training process, the peak memory usage is about 60 GiB (differing among cities).

\begin{table}[hbtp]
\centering
\caption{Hyperparameters of the LightGBM model.}
\label{tab:lgb-param}
\begin{tabular}{cr} 
\toprule
Parameter & Value \\ 
\midrule
Number of trees & $2000$ \\
Number of leaves & $64$ \\
Maximum tree depth & $7$ \\
Feature subsampling rate & $0.7$ \\
Learning rate & $0.1$ \\
\bottomrule
\end{tabular}
\end{table}

\section{Feature Importance}

\begin{table}[hbtp]
\centering
\caption{The top-20 most important features of the London model.}
\label{tab:feature-importance}
\begin{tabular}{llr}
    \toprule
    \textbf{ Feature } & \textbf{ Group } & \textbf{ Relative Importance } \\ 
    \midrule
    Median of $y$\ ($50$-NN)  & Similarity-based & 56.6\% \\ 
    Median of $y$\ ($30$-NN)  & Similarity-based & 11.9\% \\ 
    25-th percentile of $y$\ ($10$-NN)  & Similarity-based & 4.1\% \\ 
    Median of $y$\ ($30$-NN, normalized Euclidean) & Similarity-based & 3.7\% \\
    25-th percentile of $y$\ ($30$-NN)  & Similarity-based & 3.4\% \\ 
    25-th percentile of $y$\ ($50$-NN)  & Similarity-based & 2.9\% \\ 
    75-th percentile of $y$\ ($50$-NN)  & Similarity-based & 2.6\% \\ 
    Median of $y$\ ($100$-NN)  & Similarity-based & 2.5\% \\ 
    75-th percentile of $y$\ ($100$-NN)  & Similarity-based & 2.1\% \\ 
    75-th percentile of $y$\ ($30$-NN)  & Similarity-based & 1.9\% \\ 
    Median of $y$\ ($10$-NN)  & Similarity-based & 1.8\% \\ 
    Median of $y$\ ($50$-NN, normalized Euclidean) & Similarity-based & 1.5\% \\
    Supersegment ID  & Static & 0.9\% \\ 
    75-th percentile of $y$\ ($30$-NN, normalized Euclidean) & Similarity-based & 0.3\% \\
    Median of $y$\ ($50$-NN) $-$ shortest travel time & Combined & 0.3\% \\
    Median of $y$\ ($50$-NN) $-$ standard deviation of $y$\ ($50$-NN) & Combined & 0.2\% \\
    25-th percentile of $y$\ ($5$-NN)  & Similarity-based & 0.2\% \\ 
    75-th percentile of $y$\ ($10$-NN)  & Similarity-based & 0.2\% \\ 
    25-th percentile of $y$\ ($100$-NN)  & Similarity-based & 0.2\% \\ 
    Median of $y$\ ($50$-NN) $+2\times$ standard deviation of $y$\ ($50$-NN) & Combined & 0.1\% \\
    \bottomrule
\end{tabular}
\end{table}

\end{document}